\documentclass{article}
\usepackage[final]{nips_2018}

\usepackage[utf8]{inputenc}
\usepackage[T1]{fontenc}
\usepackage[
        activate={true,nocompatibility},
        tracking=true
        kerning=true,
        spacing=true
]{microtype}
\usepackage{hanging}

\usepackage{amsthm}
\usepackage{amssymb}
\usepackage{amsmath}
\usepackage{amsfonts}
\usepackage{mathtools}
\usepackage{nicefrac}
\usepackage{siunitx}

\usepackage{algorithm}
\usepackage[noend]{algpseudocode}

\usepackage{booktabs}
\usepackage{multirow}
\usepackage{graphicx}
\usepackage{subfig}

\usepackage{natbib}
\usepackage{hyperref}
\usepackage{url}

\newcommand{\dom}{\operatorname{dom}}
\newcommand{\reals}{\mathbb{R}}
\DeclarePairedDelimiter\abs{\lvert}{\rvert}

\newcommand{\E}{\operatorname{E}}

\newcommand{\ind}[1]{1\{#1\}}

\newcommand{\Sp}{\operatorname{Sp}}
\newcommand{\tr}{\operatorname{tr}}
\newcommand{\eigmin}{\lambda_\text{min}}
\newcommand{\eigmax}{\lambda_\text{max}}

\DeclarePairedDelimiterX{\inner}[2]{\langle}{\rangle}{#1, #2}
\DeclarePairedDelimiterX{\norm}[1]{\|}{\|}{#1}

\newcommand{\sci}[1]{\num[round-mode=places,round-precision=2,scientific-notation=true]{#1}}
\newcolumntype{N}{S[round-mode=places,round-precision=2]}

\newcommand{\kl}{d_\text{KL}}
\newcommand{\effFac}{v_\text{eff}}

\title{Backpropagation for Implicit Spectral Densities}
\author{%
	\textbf{Aditya Ramesh} \qquad \textbf{Yann LeCun} \\
	Department of Computer Science, New York University \\
	\texttt{\{ar2922,yann\}@cs.nyu.edu}
}

\begin{document}
\maketitle

\begin{abstract}
Most successful machine intelligence systems rely on gradient-based learning, which is made possible
by backpropagation. Some systems are designed to aid us in interpreting data when explicit goals
cannot be provided. These unsupervised systems are commonly trained by backpropagating through a
likelihood function. We introduce a tool that allows us to do this even when the likelihood is not
explicitly set, by instead using the \emph{implicit} likelihood of the model. Explicitly defining
the likelihood often entails making heavy-handed assumptions that impede our ability to solve
challenging tasks. On the other hand, the implicit likelihood of the model is accessible without the
need for such assumptions. Our tool, which we call \emph{spectral backpropagation}, allows us to
optimize it in much greater generality than what has been attempted before. GANs can also be viewed
as a technique for optimizing implicit likelihoods. We study them using spectral backpropagation in
order to demonstrate robustness for high-dimensional problems, and identify two novel properties of
the generator~$G$: (1)~there exist aberrant, nonsensical outputs to which~$G$ assigns very high
likelihood, and (2)~the eigenvectors of the metric induced by~$G$ over latent space correspond to
quasi-disentangled explanatory factors.
\end{abstract}

\section{Introduction}\label{sec:intro}

Density estimation is an important component of unsupervised learning. In a typical scenario, we are
given a finite sample~$D \coloneqq \{x_1, \ldots, x_n\}$, and must make decisions based on
information about the hypothetical process that generated~$D$. Suppose that~$D \sim P$, where~$P$ is
a distribution over some sample space~$X$. We can model this generating process by learning a
probabilistic model~$f : Z \to X$ that transforms a simple distribution~$P_Z$ over a latent
space~$Z$ into a distribution~$Q$ over~$X$. When~$Q$ is a good approximation to~$P$, we can use it
to generate samples and perform inference. Both operations are integral to the design of intelligent
systems.

One common approach is to use gradient-based learning to maximize the likelihood of~$D$ under the
model distribution~$Q$. We typically place two important constraints on~$f$ in order to make this
possible: (1)~existence of an explicit inverse~$f^{-1} : X \to Z$, and~(2) existence of a simple
procedure by which we can evaluate~$Q$. Often times, $f$~is instead constructed as a map from~$X$
to~$Z$ in order to make it convenient to evaluate~$Q(x)$ for a given observation~$x \in X$. This is
the operation on which we place the greatest demand for throughput during training. Each choice
corresponds to making one of the two operations -- generating samples or evaluating~$Q$ --
convenient, and the other inconvenient. Regardless of the choice, both operations are required, and
so both constraints are made to hold in practice. We regard~$f$ as a map from~$Z$ to~$X$ in this
presentation for sake of simplicity.

Much of the work in deep probabilistic modeling is concerned with allowing~$f$ to be flexible enough
to capture intricate latent structure, while simultaneously ensuring that both conditions hold. We
can dichotomize current approaches based on how the second constraint -- existence of a simple
procedure to evaluate~$Q$ -- is satisfied. The first approach involves making~$f$ autoregressive by
appropriately masking the weights of each layer. This induces a lower-triangular structure in the
Jacobian, since each component of the model's output is made to depend only on the previous ones. We
can then rapidly evaluate the log-determinant term involved in likelihood computation, by
accumulating the diagonal elements of the Jacobian.

Research into autoregressive modeling dates back several decades, and we only note some recent
developments. \cite{MADE} describe an autoregressive autoencoder for density estimation. \cite{IAF}
synthesize autogressive modeling with normalizing flows~\citep{NF} for variational inference, and
\cite{MAF} make further improvements. \cite{PixelRNN} apply this idea to image generation, with
follow-up work~(\cite{RealNVP}, \cite{PixelCNN}) that exploits parallelism using masked
convolutions. \cite{WaveNet} do the same for audio generation, and \cite{ParallelWaveNet} introduce
strategies to improve efficiency. We refer the reader to~\cite{NFTutorial} for an excellent overview
of these works in more detail.

The second approach involves choosing the layers of~$f$ to be transformations, not necessarily
autoregressive, for which explicit expressions for the Jacobian are still available. We can then
evaluate the log-determinant term for~$f$ by accumulating the layerwise contributions in accordance
with the chain rule, using a procedure analogous to backpropagation. \cite{NF} introduced this idea
to variational inference, and recent work, including~\citep{SylvesterNF} and \citep{HouseholderNF},
describe new types of such transformations.

Both approaches must invariably compromise on model flexibility. An efficient method for
differentiating implicit densities that do not fulfill these constraints would enrich the current
toolset for probabilistic modeling. \cite{AISEval}~advocate using annealed importance
sampling~\citep{AIS} for evaluating implicit densities, but it is not clear how this approach could
be used to obtain gradients. Very recent work~\citep{SteinGradEst} uses Stein's identity to cast
gradient computation for implicit densities as a sparse recovery problem. Our approach, which we
call \emph{spectral backpropagation}, harnesses the capabilities of modern automatic
differentiation~(\cite{TF}, \cite{PyTorch}) by directly backpropagating through an approximation for
the spectral density of~$f$.

We make the first steps toward demonstrating the viability of this approach by minimizing~$\kl(Q,
P_X)$ and~$\kl(P_X, Q)$, where~$Q$ is the implicit density of a non-invertible
Wide~ResNet~\citep{WideResNet}~$f$, on a set of test problems. Having done so, we then turn our
attention to characterizing the behavior of the generator~$G$ in GANs~\citep{GAN}, using a series of
computational studies made possible by spectral backpropagation. Our purpose in conducting these
studies is twofold. Firstly, we show that our approach is suitable for application to
high-dimensional problems. Secondly, we identify two novel properties of generators:

\begin{itemize}
\item The existence of adversarial perturbations for classification models~\citep{NNProperties} is
paralleled by the existence of aberrant, nonsensical outputs to which~$G$ assigns very high
likelihood.
\item The eigenvectors of the metric induced by the~$G$ over latent space correspond to meaningful,
quasi-disentangled explanatory factors. Perturbing latent variables along these eigenvectors allows
us to quantify the extent to which~$G$ makes use of latent space.
\end{itemize}

We hope that these observations will contribute to an improved understanding of how well generators
are able to capture the latent structure of the underlying data-generating process.

\section{Background}

\subsection{Generalizing the Change of Variable Theorem}

We begin by revisiting the geometric intuition behind the usual change of variable theorem. First,
we consider a rectangle in~$\reals^2$ with vertices~$x_0, x_1, x_3, x_2$ given in clockwise order,
starting from the bottom-left vertex. To determine its area, we compute its side lengths~$v_1
\coloneqq x_1 - x_0, v_2 \coloneqq x_2 - x_0$ and write~$V_2 = v_1 v_2$. Now suppose we are given a
parallelepiped in~$\reals^3$ whose sides are described by the vectors~$v_1 \coloneqq x_1 - x_0$,
$v_2 \coloneqq x_2 - x_0$, and $v_3 \coloneqq x_3 - x_0$. Its volume is given by the triple product
$V_3 = \inner{v_1 \times v_2}{v_3}$, where~$\times$ and~$\inner{\cdot}{\cdot}$ denote cross product
and inner product, respectively. This triple product can be rewritten as
\[
	V_3 = \det\begin{pmatrix} v_1 & v_2 & v_3 \end{pmatrix},
\]
which we can generalize to compute the volume of a parallelepiped in~$\reals^N$:
\[
	V_N = \det\begin{pmatrix} v_1 & \cdots & v_N \end{pmatrix}.
\]
If we regard the vertices~$x_0, \ldots, x_N$ as observations in~$X$, the change of variable theorem
can be understood as the differential analog of this formula. To wit, we suppose that~$f: \reals^N
\to \reals^N$ is a diffeomorphism, and denote by~$J_f$ the Jacobian of its output with respect to
its input. Now,~$V_N$ becomes the infinitesimal volume element determined by~$J_f$. For an
observation~$x \in X$, the change of variable theorem says that we can compute
\[
	Q(x) = P_Z(f^{-1}(x)) \,\abs{\det J_f(f^{-1}(x)))}^{-1}
	\eqqcolon P_Z(z) \,\abs{\det(J_f(z))}^{-1},
\]
where we set~$z \coloneqq f^{-1}(x)$.

An $n$-dimensional parallelepiped in~$\reals^N$ requires~$n$ vectors to specify its sides. When~$n
< N$, its volume is given by the more general formula~\citep{NDimGraphics},
\[
	V_n^2 = \det\begin{pmatrix}
	\inner{v_1}{v_1} & \cdots & \inner{v_1}{v_n} \\
	\vdots           & \ddots & \vdots \\
	\inner{v_n}{v_1} & \cdots & \inner{v_n}{v_n}
	\end{pmatrix}.
\]
The corresponding analog of the change of variable theorem is known in the context of geometric
measure theory as the smooth coarea formula~\citep{Coarea}. When~$f$ is a diffeomorphism between
manifolds, it says that
\begin{equation}
	\label{eq:coarea}
	Q(x) = P_Z(f^{-1}(x)) \det(J_f(f^{-1}(x))^t J_f(f^{-1}(x)))^{-1 / 2}
	\eqqcolon P_Z(z) \det{M_f(z)}^{-1 / 2},
\end{equation}
where we set~$z \coloneqq f^{-1}(x)$ as before, and define~$M_f \coloneqq (J_f)^t J_f$ to be the
metric induced by~$f$ over the latent manifold~$Z$. In many cases of interest, such as in GANs, the
function~$f$ is not necessarily injective.  Application of the coarea formula would then require us
to evaluate an inner integral over~$\{z \in Z : f(z) = x\}$, rather than over the
singleton~$\{f^{-1}(x)\}$. We ignore this technicality and apply Equation~\ref{eq:coarea} anyway.

The change of variable theorem gives us access to the implicit density~$Q$ in the form of the
\emph{spectral density} of~$M_f$. Indeed, the Lie identity~$\ln \det = \tr \ln$ allows us to express
the log-likelihood corresponding to Equation~\ref{eq:coarea} as
\begin{equation}
	\label{eq:ll}
	\ln Q(x) = \ln P(z) - \frac{1}{2} \ln \det M_f(z) = \ln P(z) - \frac{1}{2} \tr \ln M_f(z).
\end{equation}
We focus on the factor involving~$M_f$ on the~RHS, which can be written as
\[
	\tr \ln M_f(z) = \sum_{\lambda \in \Sp(M_f(z))} \ln \lambda = \E_{\lambda \sim P_\lambda} \ln \lambda,
\]
where~$\Sp$ denotes the spectrum, and~$P_\lambda$ the delta distribution over the eigenvalues in the
spectrum. We let~$\theta$ denote the parameters of~$f$, and assume that~$P_Z$ is independent
of~$\theta$. Now, differentiating Equation~\ref{eq:ll} with respect to~$\theta$ gives
\begin{equation}
	\label{eq:spectral_bprop}
	D_\theta \ln Q(x) = -D_\theta \E_{\lambda \sim P_\lambda} \ln \lambda.
\end{equation}
Equation~\ref{eq:spectral_bprop} allows us to formulate gradient computation for implicit densities
as a variant of stochastic backpropagation~(\cite{AEVB}, \cite{StochasticBackprop}), in which the
base distribution for the expectation is the spectral density of~$M_f$ rather than a normal
distribution.

\subsection{An Estimator for Spectral Backpropagation}

\begin{algorithm}[t]
\caption{Procedure to estimate~$\ln \det$ using the Chebyshev
approximation.\label{alg:log_det_chebyshev}}

\begin{algorithmic}
\Require{$A \in \reals^{n \times n}$ is the implicit matrix;~$m$ the desired order;~$p$ the number
of probe vectors for the trace estimator;~$t$ the number of power iterations;~$g \geq 1$ a
multiplier for the estimate returned by the power method; and~$\epsilon$ the stipulated lower bound
on~$\Sp(A)$.}

\Procedure{StochasticLogDet}{$A, m, p, t, g, \epsilon$}
\State $\hat{\lambda}_\text{max} \gets \Call{PowerMethod}{A, t}$
\State $\mu, \nu \gets \epsilon, g \,\hat{\lambda}_\text{max}$
\State $a, b \gets \mu / (\mu + \nu), \nu / (\mu + \nu)$
\State Define~$\varphi$ and~$\varphi^{-1}$ using Equation~\ref{eq:rescale_func}.
\State $\{c_i\}_{i \in [0, m]} \gets \Call{ChebyshevCoefficients}{\ln \circ\, \varphi}$
\State $\bar{A} \gets A / (\mu + \nu)$
\State $\Gamma \gets \Call{StochasticChebyshevTrace}{\varphi^{-1}(\bar{A}), \{c_i\}, p}$
\State \Return $n\ln(a + b) + \Gamma$
\EndProcedure

\Procedure{StochasticChebyshevTrace}{$A, \{c_i\}_{i \in [0, m]}, p$}
\State $r \gets 0$
\For{$j \in [1, p]$}
	\State $v \gets \Call{RandomRademacher}{n}$
	\State $w_0, w_1 \gets v, Av$
	\State $s \gets c_0 w_0 + c_1 w_1$

	\For{$i \in [2, m]$}
		\State $w_i \gets 2 A w_{i - 1} - w_{i - 2}$
		\State $s \gets s + c_i w_i$
	\EndFor

	\State $r \gets r + \inner{v}{s}$
\EndFor
\State \Return $r / p$
\EndProcedure
\end{algorithmic}
\end{algorithm}

To obtain an estimator for Equation~\ref{eq:spectral_bprop}, we turn to the thriving literature on
stochastic approximation of spectral sums. These methods estimate quantities of the
form~$\Sigma_S(A) \coloneqq \tr S(A)$, where~$A$ is a large or implicitly-defined matrix, by
accessing~$A$ using only matrix-vector products. In our case,~$S = \ln$, and the products
involving~$A = M_f$ can be evaluated rapidly using automatic differentiation. We make no attempt to
conduct a comprehensive survey, but note that among the most promising recent approaches are those
described by~\cite{SSChebyshev}, \cite{SSTaylor}, \cite{SSLanczos}, and~\cite{SSEntropic}.

We briefly describe the approaches of~\cite{SSChebyshev} and~\cite{SSTaylor}, which work on the
basis of polynomial interpolation. Given a function~$\bar{S}: [-1, 1] \to \reals$, these methods
construct an order-$m$ approximating polynomial~$\bar{p}_m$ to~$\bar{S}$, given by
\[
	\bar{S} \approx \bar{p}_m \coloneqq \sum_{i \in [0, m]} c_i T_i,
\]
where~$c_i \in \reals$ and~$T_i : [-1, 1] \to \reals$. The main difference between the two
approaches is the choice of approximating polynomial. \cite{SSTaylor} use Taylor polynomials, for
which
\[
	c_i \coloneqq \frac{\bar{S}^{(i)}}{i!} \quad\text{and}\quad
	T_i : x \mapsto x^i,
\]
where we use superscript~$(i)$ to denote iterated differentiation. On the other
hand, \cite{SSChebyshev} use Chebyshev polynomials. These are defined by the recurrence relation
\begin{equation}
	\label{eq:chebyshev}
	T_0 = 1, \quad T_1 : x \mapsto x, \quad\text{and}\quad
	T_i : x \mapsto 2x T_{i - 1}(x) - T_{i - 2}(x), \quad i \geq 2.
\end{equation}
The coefficients~$\{c_i\}$ for the Chebyshev polynomials are called the \emph{Chebyshev nodes}, and
are defined by
\[
	c_i \coloneqq \begin{dcases}
	\frac{1}{m + 1} \sum_{j \in [0, m]} \bar{S}(x_j) T_0(x_j), &i = 0, \\
	\frac{2}{m + 1} \sum_{j \in [0, m]} \bar{S}(x_j) T_i(x_j), &i \geq 1.
	\end{dcases}
\]
Now suppose that we are given a matrix~$\bar{A} \in \reals^{n \times n}$ such that~$\Sp(\bar{A})
\subset [-1, 1]$. After having made a choice for the construction of~$\bar{p}_m$, we can use the
approximation
\begin{align}
	\Sigma_{\bar{S}}(\bar{A})
	&= \tr \bar{S}(\bar{A})
	= \sum_{\lambda \in \Sp(\bar{A})} \bar{S}(\lambda)
	\approx \sum_{\lambda \in \Sp(\bar{A})} \bar{p}_m(\lambda) \nonumber\\
	&= \sum_{\lambda \in \Sp(\bar{A})} \sum_{i \in [0, m]} c_i T_i(\lambda)
	= \sum_{i \in [0, m]} c_i \sum_{\lambda \in \Sp(\bar{A})} T_i(\lambda) \nonumber\\
	&= \sum_{i \in [0, m]} c_i \tr T_i(\bar{A}). \label{eq:ss_traces}
\end{align}
This reduces the problem of estimating the spectral sum~$\Sigma_{\bar{S}}(\bar{A})$ to computing the
traces~$\tr T_i(\bar{A})$ for all~$i \in [0, m]$.

Two issues remain in applying this approximation. The first is that both~$\dom(\bar{S})$
and~$\Sp(\bar{A})$ are restricted to~$[-1, 1]$. In our case, $\ln : (0, \infty) \to \reals$, and
$M_f(z)$ can be an arbitrary positive definite matrix. To address this issue, we define~$\varphi:
[-1, 1] \to [a, b]$, where
\begin{equation}
	\label{eq:rescale_func}	
	\varphi: x \mapsto \frac{b - a}{2} x + \frac{b + a}{2} \quad\text{and}\quad
	\varphi^{-1}: x \mapsto \frac{2}{b - a} x - \frac{b + a}{b - a}.
\end{equation}
Now we set~$\bar{S} \coloneqq S \circ \varphi$, so that
\[
	S = \bar{S} \circ \varphi^{-1} \approx \bar{p}_m \circ \varphi^{-1}
	= \sum_{i \in [0, m]} c_i (T_i \circ \varphi^{-1}) \eqqcolon p_m.
\]
We stress that while~$p_m$ is defined using~$\varphi^{-1}$, the coefficients~$c_i$ are computed
using~$\bar{S} \coloneqq S \circ \varphi$. With these definitions in hand, we can write
\begin{equation}
	\label{eq:sss}
	\Sigma_S(\bar{A}) = \Sigma_{\bar{S} \circ \varphi^{-1}}(\bar{A}) \approx \tr p_m(\bar{A}).
\end{equation}
\cite{SSChebyshev} require spectral bounds~$\mu$ and~$\nu$, so that~$\Sp(A) \subset [\mu, \nu]$, and
set
\[
	a \coloneqq \frac{\mu}{\mu + \nu}, \quad
	b \coloneqq \frac{\nu}{\mu + \nu}, \quad\text{and}\quad
	\bar{A} \coloneqq \frac{A}{a + b}.
\]
After using Equation~\ref{eq:sss} with a Chebyshev approximation for~$S = \ln$ to obtain~$\Gamma
\approx \ln \det(\bar{A})$, we compute
\[
	\ln \det(A) = n \ln(a + b) + \ln \det(\bar{A}) \approx n \ln(a + b) + \Gamma.
\]
\cite{SSTaylor} instead define~$B \coloneqq A / \nu$, and write
\[
	\ln \det(B) = \tr \ln(B) = \tr \ln(I - (I - B)).
\]
This time, we set~$\bar{A} = I - B$ and use Equation~\ref{eq:sss} with a Taylor approximation for~$S
= \ln(1 - x)$ to obtain~$\Gamma \approx \ln \det(B)$. Then, we compute
\[
	\ln \det(A) = n \ln \nu + \ln \det(\bar{A}) \approx n \ln \nu + \Gamma.
\]
We can easily obtain an accurate upper bound~$\nu$ using the power method. The lower bound~$\mu$ is
fixed to a small, predetermined constant in our work.

The second issue is that deterministically evaluating the terms~$\tr T_i(\bar{A})$ in
Equation~\ref{eq:ss_traces} requires us to compute matrix powers of~$\bar{A}$. Thankfully, we can
drastically reduce the computational cost and approximate these terms using only matrix-vector
products. This is made possible by the stochastic trace estimator introduced by~\cite{TraceEst}:
\[
	\tr \bar{A} = \E_{v \sim P_V} \inner{v}{\bar{A}v}
	\approx \frac{1}{p} \sum_{j \in [1, p]} \inner{v_j}{\bar{A} v_j}.
\]
When the distribution~$P_V$ for the probe vectors~$v$ has expectation zero, the estimate is
unbiased. We use the Rademacher distribution, which samples the components of~$v$ uniformly
from~$\{-1, 1\}$. We refer the reader to~\cite{TraceEstVariance} for a detailed study on the
variance of this estimator.

We first describe how the trace estimator is applied when a Taylor approximation is used to
construct~$\bar{p}_m$. In this case, we have
\[
	\tr \bar{p}_m(\bar{A})
	= \sum_{i \in [0, m]} c_i \tr \bar{A}^i
	\approx \frac{1}{p} \sum_{i \in [0, m]} c_i \sum_{j \in [1, p]} \inner{v_j}{\bar{A}^i v_j}
	= \frac{1}{p} \sum_{j \in [1, p]} c_i \left\langle v_j, \sum_{i \in [0, m]} \bar{A}^i v_j \right\rangle.
\]
The inner summands~$\bar{A}^i v_j$ are evaluated using the recursion~$w_0 \coloneqq v_j$ and~$w_i
\coloneqq \bar{A} w_{i - 1}$ for~$i \geq 1$. It follows that the number of matrix-vector products
involved in the approximation increases linearly with respect to the order~$m$ of the approximating
polynomial~$\bar{p}_m$. The same idea allows us to accumulate the traces for the Chebyshev
approximation, based on Equation~\ref{eq:chebyshev}. The resulting procedure is given in
Algorithm~\ref{alg:log_det_chebyshev}; it is our computational workhorse for evaluating the
log-likelihood in Equation~\ref{eq:ll} and estimating the gradient in
Equation~\ref{eq:spectral_bprop}.

\section{Learning Implicit Densities}

\begin{figure}[t]
\centering%
\begin{minipage}[t]{0.71\textwidth}%
\centering%
\subfloat[][Plots of the loss and relative error for minimizing~$\kl(Q, P)$.]{%
	{\def\arraystretch{0}%
	\begin{tabular}[t]{c}%
	\includegraphics[width=\textwidth]{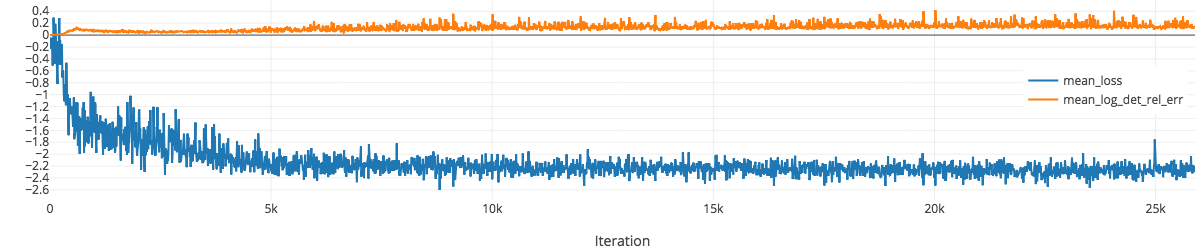} \\
	\includegraphics[width=\textwidth]{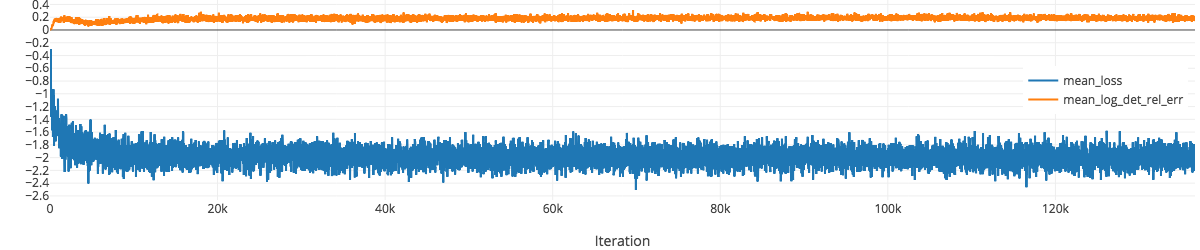} \\
	\includegraphics[width=\textwidth]{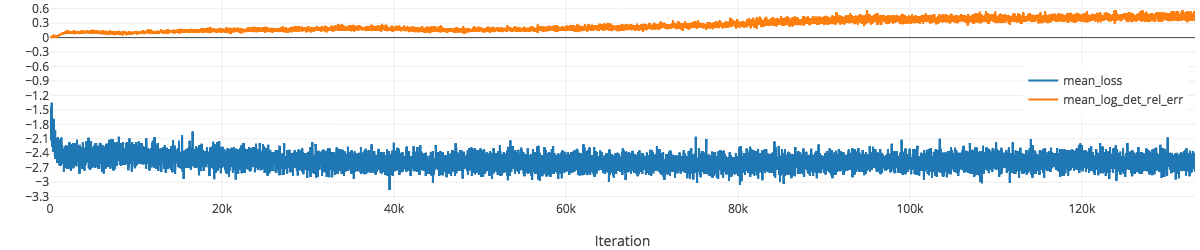} \\
	\includegraphics[width=\textwidth]{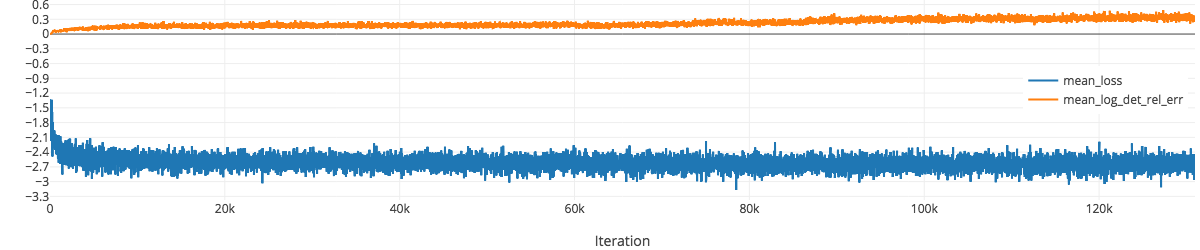}%
	\end{tabular}}%
}%
\end{minipage}\hspace{0.01\textwidth}%
\begin{minipage}[t]{0.28\textwidth}%
\centering%
{\captionsetup{width=0.9\linewidth}%
\subfloat[][Samples from~$f$ for the four test energies at epochs~5, 10, 9, and 4, respectively.]{%
	{\def\arraystretch{0}%
	\begin{tabular}[t]{c@{}c}%
	\includegraphics[width=0.45\textwidth]{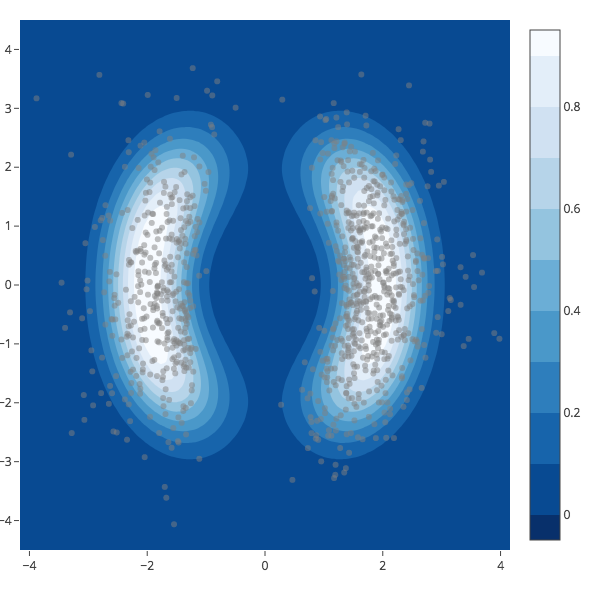} &
	\includegraphics[width=0.45\textwidth]{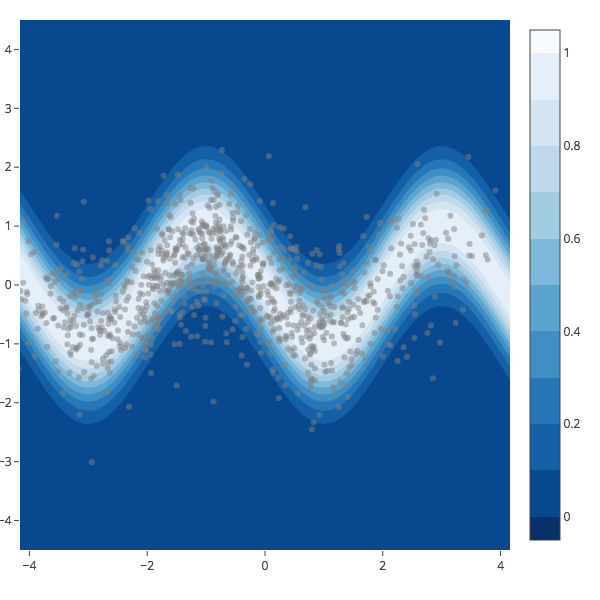} \\
	\includegraphics[width=0.45\textwidth]{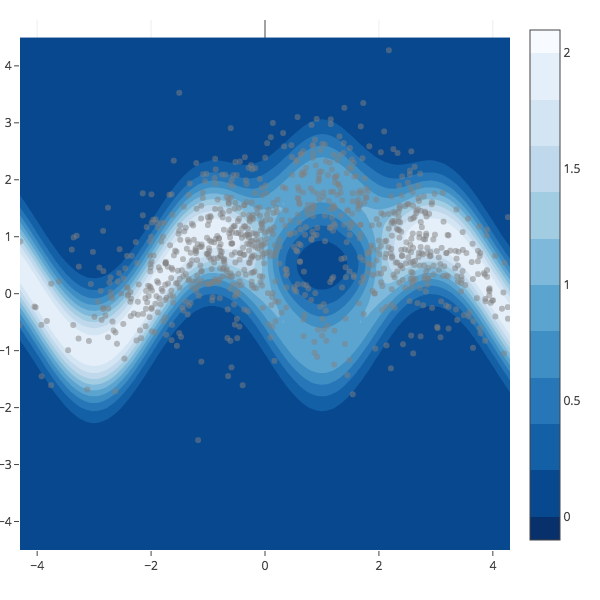} &
	\includegraphics[width=0.45\textwidth]{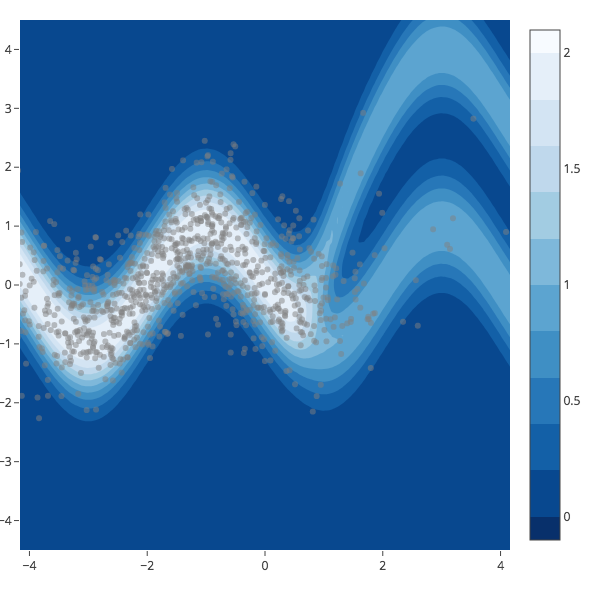}
	\end{tabular}}%
}} \\
{\captionsetup{width=0.9\linewidth}%
\subfloat[][Samples from~$f$ for the last two test energies, both at epoch~26.]{%
	{\def\arraystretch{0}%
	\begin{tabular}[b]{c@{}c}%
	\includegraphics[width=0.45\textwidth]{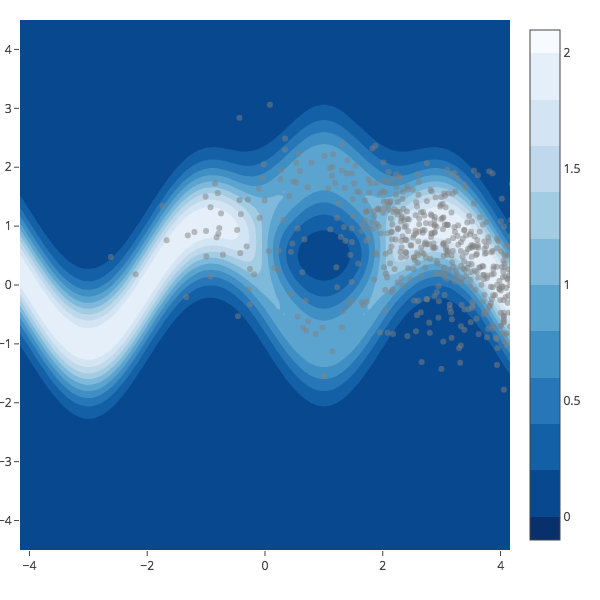} &
	\includegraphics[width=0.45\textwidth]{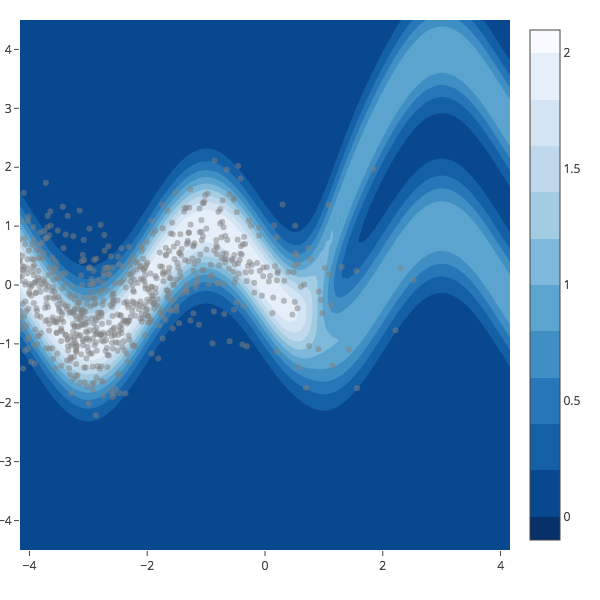}
	\end{tabular}}%
}}
\end{minipage}%
\caption{Results for minimizing~$\kl(Q, P_X)$ for the four test energies described in~\cite{NF}.
Subfigures~(b) and~(c) show the model samples superimposed over contour plots of the corresponding
ground-truth test energies. Each epoch corresponds to \num{5000} iterations. We see in~(a) that the
relative error for the approximation to the log determinant typically stays below~30\%, except
toward the end of training for the last two test energies. At this point, samples from these two
models begin to drift away from the origin, as shown in~(c).\label{fig:reverse_kl_results}}
\end{figure}

Suppose that we are tasked with matching a given data distribution~$P_X$ with the implicit
density~$Q$ of the model~$f$. Two approaches for learning~$f$ are available, and the choice of which
to use depends on the type of access we have to the data distribution~$P_X$. The first approach --
minimizing $\kl(Q, P_X)$ -- is applicable when we know how to evaluate the likelihood of~$P_X$, but
are not necessarily able to sample from it. The second approach -- minimizing $\kl(P_X, Q)$ -- is
applicable when we are able to sample from~$P_X$, but are not necessarily able to evaluate its
likelihood. We show that spectral backpropagation can be used in both cases, when neither of the two
conditions described in Section~\ref{sec:intro} holds.

All of the examples considered here are densities over~$\reals^2$. We match them by transforming a
prior~$P_Z$ given by a spherical normal distribution. Our choice for the architecture of~$f$ is a
Wide ResNet comprised of four residual blocks. Each block is a three-layer bottleneck
from~$\reals^2$ to~$\reals^2$ whose hidden layer size is~32. All layers are equipped with biases,
and use LeakyReLU activations. We compute the gradient updates using a batch size of~64, and apply
the updates using Adam~\citep{Adam} with a step size of~$\sci{1e-4}$, and all other parameters kept
at their the default values.

To compute the gradient update given by Equation~\ref{eq:spectral_bprop}, we use
Algorithm~\ref{alg:log_det_chebyshev} with~$(m, p, t, g) \coloneqq (10, 20, 20, 1.2)$ for all
experiments. For minimizing~$\kl(Q, P_X)$, we use~$\epsilon \coloneqq 0.1$, and for
minimizing~$\kl(P_X, Q)$, $\epsilon \coloneqq \sci{1e-2}$. In order to monitor the accuracy of the
approximation for the likelihood, we compute~$M_f$ at each iteration, and evaluate the ground-truth
likelihood in accordance with Equation~\ref{eq:ll}. We define the relative error of the
approximation~$\ln \hat{\ell}$ with respect to the ground-truth log-likelihood~$\ln \ell$ by
\[
	\ln \hat{\ell} - \ln \ell
	= \ln \frac{\hat{\ell}}{\ell}
	= \ln\left(1 + \frac{\hat{\ell} - \ell}{\ell}\right)
	\approx \frac{\hat{\ell} - \ell}{\ell},
\]
provided that the quotient is not too large. This definition of relative error avoids numerical
problems when~$\ell \approx 0$.

We begin by considering the first approach, in which we seek to minimize~$\kl(Q, P_X)$. This
objective requires that we be able to sample from~$Q$, so we choose~$f: Z \to X$ to be a map from
latent space to observation space. The results are shown in Figure~\ref{fig:reverse_kl_results}. To
prevent~$f$ from making~$Q$ collapse to model the infinite support of~$P_X$, we found it helpful to
incorporate the regularizer
\begin{equation}
	\label{eq:jacobian_penalty}
	R(M_f) \coloneqq \rho \norm{M_f}_2
\end{equation}
into the objectives for the third and fourth test energies. Here,~$\norm{\cdot}_2$ denotes the
spectral norm. To implement this regularizer, we simply backpropagate through the estimate
of~$\eigmax$ that is already produced by Algorithm~\ref{alg:log_det_chebyshev}. We use~$\rho
\coloneqq \sci{8e-2}$ in both cases. Despite the use of this regularizer, we find that continuing to
train the models for these last two test energies causes the samples to drift away from the
origin~(see Figure~\ref{fig:reverse_kl_results}(c)). We have not made any attempt to address this
behavior.  Finally, we note that since~$\kl(Q, P_X)$ is bounded from below by the negative
log-normalization constant of~$Q$, it can become negative when~$Q$ is unnormalized. We see that this
happens for all four examples.

\begin{figure}[t]
\centering
\begin{minipage}[t]{0.71\textwidth}
\centering%
\subfloat[][Plots for the loss and relative error for minimizing~$\kl(P, Q)$.]{%
	{\def\arraystretch{0}%
	\begin{tabular}[t]{c}%
	\includegraphics[width=\textwidth]{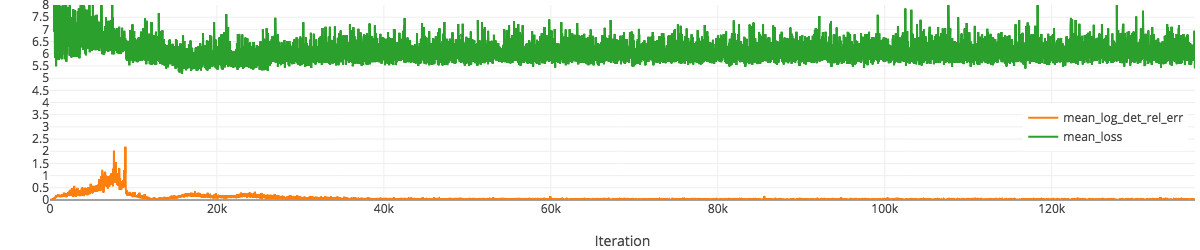} \\
	\includegraphics[width=\textwidth]{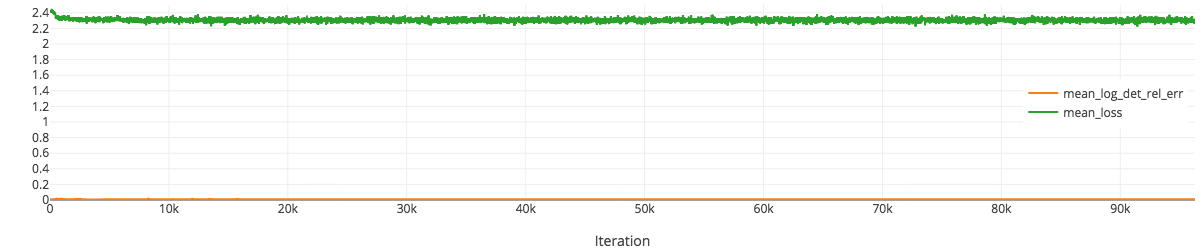}
	\end{tabular}}
}
\end{minipage}\hspace{0.02\textwidth}%
\begin{minipage}[t]{0.26\textwidth}%
\centering%
\subfloat[][Contour plots of the log-likelihood of~$f$ for the two test distributions, at the end of
training.]{%
	\centering%
	\includegraphics[width=0.5\textwidth]{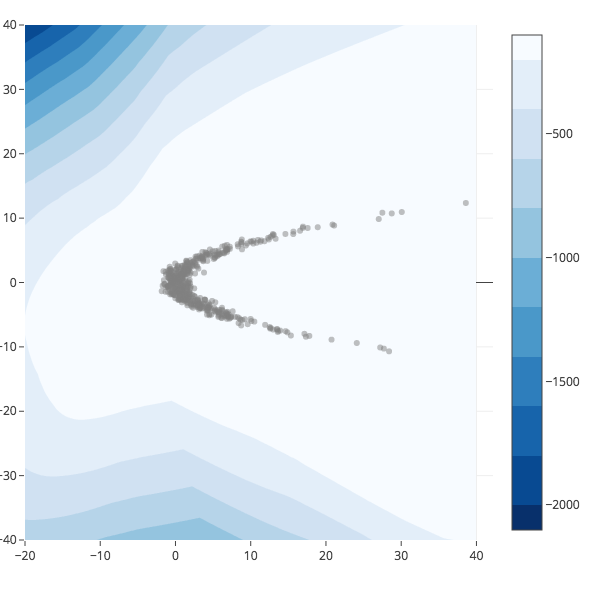}
	\includegraphics[width=0.5\textwidth]{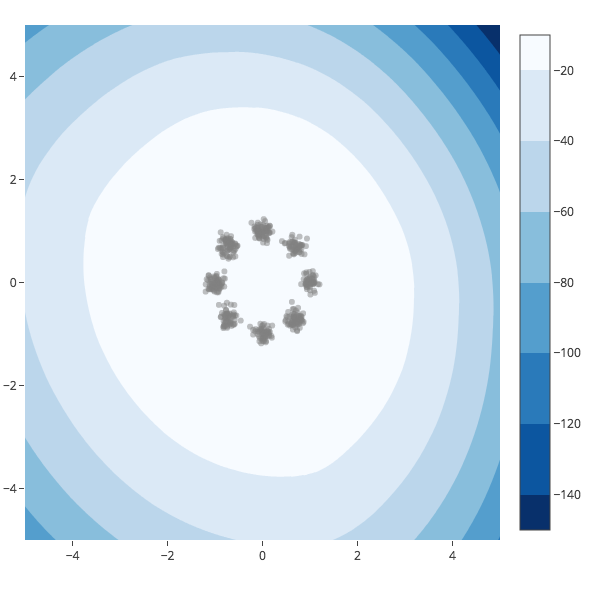}
}
\end{minipage}
\caption{Results for minimizing~$\kl(P_X, Q)$ for the crescent and circular mixture densities whose
definitions are given in the supplementary material. The relative error for the approximation to the
log determinant typically stays below~5\%. In~(b), we show samples from~$P_X$ superimposed over
contour plots of the log-likelihood of~$f$.\label{fig:forward_kl_results}}
\end{figure}

In the second approach, we seek to minimize~$\kl(P_X, Q)$. This objective requires that we be able
to evaluate the likelihood of~$Q$, so we choose~$f : X \to Z$ to be a map from observation space to
latent space. The results are shown in Figure~\ref{fig:forward_kl_results}. We note that
minimizing~$\kl(P_X, Q)$ is ill-posed when~$Q$ is unnormalized. In this scenario, the model
distribution~$Q$ can match~$P_X$ while also assigning mass outside the support of~$P_X$.  We see
that this expected behavior manifests in both examples.

\section{Evaluating GAN Likelihoods}

\begin{figure}[t]
\centering
\begin{minipage}[t]{0.7\textwidth}%
\centering%
{\captionsetup{width=0.98\linewidth,justification=centering}%
\subfloat[][Samples from the CelebA model with~$n_z = 128, n_f = 64$ (left) and the LSUN Bedroom
model with $n_z = 256, n_f = 64$ (right).]{%
	{\def\arraystretch{0}%
	\begin{tabular}{c@{}c}%
	\includegraphics[width=0.5\textwidth]{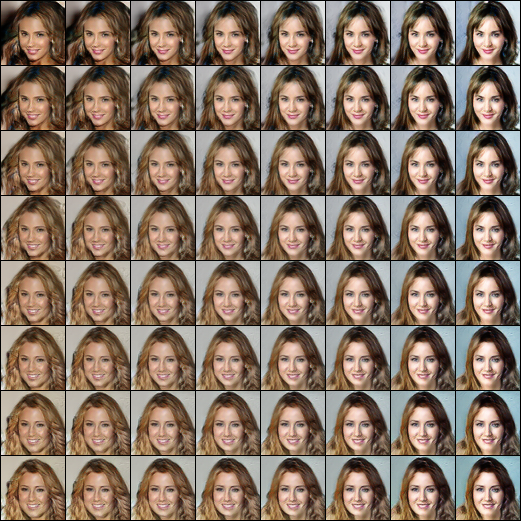} &
	\includegraphics[width=0.5\textwidth]{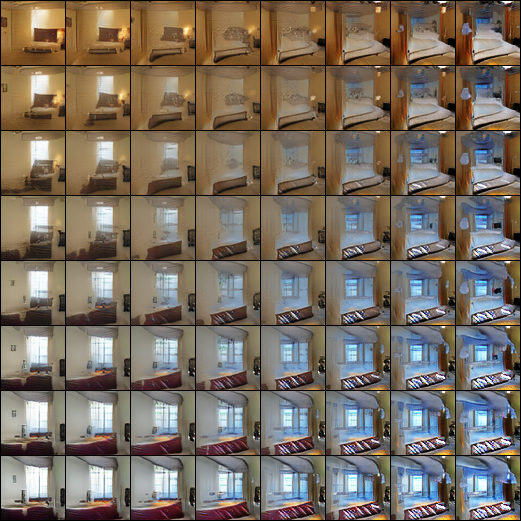}%
	\end{tabular}}%
}}%
\end{minipage}\hspace{0.01\textwidth}%
\begin{minipage}[t]{0.28\textwidth}%
\centering%
{\captionsetup{width=0.98\linewidth}%
\subfloat[][Contour plots of the log-likelihoods evaluated over the image grids to the left.]{
	{\def\arraystretch{0}%
	\begin{tabular}{c@{}c}%
	\includegraphics[width=0.5\textwidth]{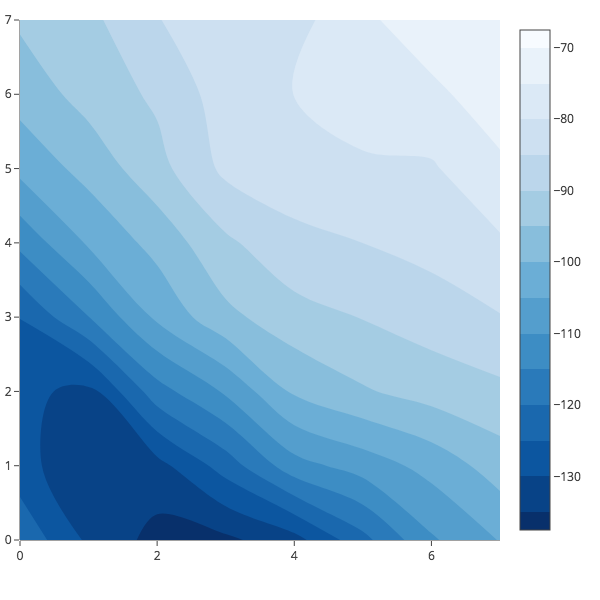} &
	\includegraphics[width=0.5\textwidth]{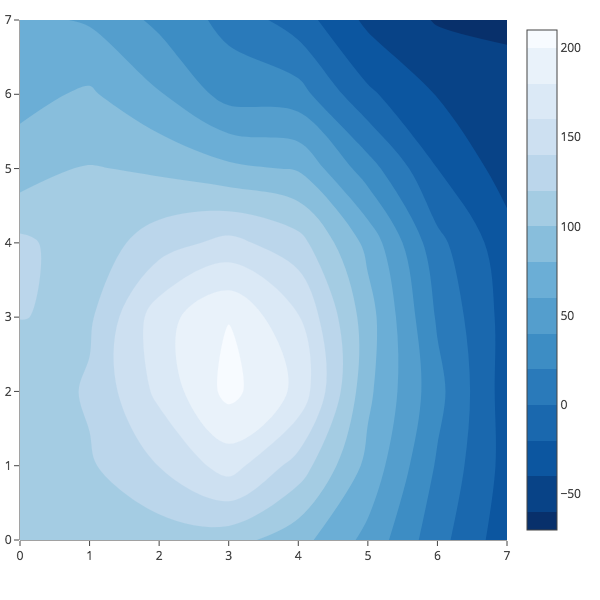}%
	\end{tabular}}%
}}%
\end{minipage}%
\caption{Samples from two models evaluated at fixed grids in latent space (left), and contour plots
of the model log-likelihood evaluated over the same grids (right).\label{fig:samples_and_contours}}
\end{figure}

For our explorations involving GANs, we train a series of DCGAN~\citep{DCGAN} models on~$64 \times
64$ rescaled versions of the CelebA~\citep{CelebA} and LSUN Bedroom datasets. We vary model capacity
in terms of the base feature map count multiplier~$n_f$ for the DCGAN architecture. The generator
and discriminator have five layers each, and use translated LeakyReLU activations~\citep{TPReLU}. To
stabilize training, we use weight normalization with fixed scale factors in the
discriminator~\citep{ImprovedGAN}. Our prior is defined by~$P_Z \coloneqq \operatorname{unif}([-1,
1])^{n_z}$, where $n_z$ is the size of the embedding space. All models were trained for
\num{750000}~iterations with a batch size of~32, using RMSProp with step size~$\sci{1e-4}$ and decay
factor~$0.9$. We present results from two of these models in Figure~\ref{fig:samples_and_contours}.

\begin{figure}
\centering
{\captionsetup{width=0.48\linewidth}%
\subfloat[][Trajectory at iterations~0, 100, 500, 1000, 2000, and 3000 (3500 total).]{%
	\includegraphics[width=0.49\textwidth]{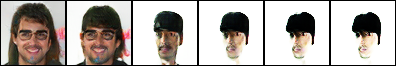}
}}%
{\captionsetup{width=0.48\linewidth}%
\subfloat[][Trajectory at iterations~0, 50, 100, 250, 500, 1000 (1000 total).]{%
	\includegraphics[width=0.49\textwidth]{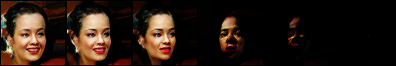}
}} \\
{\captionsetup{width=0.48\linewidth}%
\subfloat[][Trajectory at iterations~0, 100, 500, 1000, 2500, 5000 (5000 total).]{%
	\includegraphics[width=0.49\textwidth]{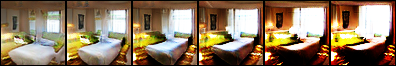}
}}%
{\captionsetup{width=0.48\linewidth}%
\subfloat[][Trajectory at iterations~0, 250, 500, 1000, 2500, 5000 (5000 total).]{%
	\includegraphics[width=0.49\textwidth]{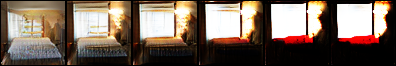}
}} \\[0.1in]
{\scriptsize%
\begin{tabular}{c N c c c}
\toprule
Trial & ${\log(p_\text{final} / p_\text{init})}$ & Initial $(\eigmin, \eigmax)$ & Final $(\eigmin, \eigmax)$ & $(m, p, t, g, \epsilon)$ \\
\midrule
(a) & 161.40078735351562 & $(\sci{0.23120332},  \sci{2907.4229})$ & $(\sci{0.008105371},   \sci{1059.2062})$  & $(5, 20, 20, 1.1, \sci{1e-4})$ \\
(b) & 667.4318695068359  & $(\sci{0.27095607},  \sci{2233.8276})$ & $(\sci{2.5146794e-06}, \sci{0.25737128})$ & $(5, 20, 20, 1.1, \sci{1e-5})$ \\
(c) & 90.82517051696777  & $(\sci{0.018747492}, \sci{1995.3479})$ & $(\sci{0.008580686},   \sci{562.2244})$   & $(5, 10, 10, 1.2, \sci{1e-2})$ \\
(d) & 336.0686254501343  & $(\sci{0.022936836}, \sci{1242.7954})$ & $(\sci{0.0006118099},  \sci{403.83206})$  & $(5, 10, 10, 1.2, \sci{1e-2})$ \\
\bottomrule
\end{tabular}}
\caption{Trajectories of four latent variables as they are perturbed to optimize likelihood under
the generator. Trajectories~(a) and~(b) correspond to the CelebA model with~$n_z = 128, n_f = 64$,
and trajectories~(c) and~(d) to the LSUN Bedroom model with~$n_z = 256, n_f = 64$. Statistics from
these trajectories are tabulated above. The last column of the table specifies the parameters used
for Algorithm~\ref{alg:log_det_chebyshev} to compute the gradient
estimates.\label{fig:ml_trajectories}}
\end{figure}

We apply spectral backpropagation to explore the effect of perturbing a given latent variable~$z \in
Z$ to maximize likelihood under the generator distribution~$Q$. This is readily accomplished by
noting that the same procedure to evaluate Equation~\ref{eq:spectral_bprop} can also be used to
obtain gradients with respect to~$z$. The results are shown in Figure~\ref{fig:ml_trajectories}.
Intuitively, we might expect the outputs to be transformed in such a way that they gravitate towards
modes of the dataset. But this is not what happens. Instead, the outputs are transformed into highly
aberrant, out-of-distribution examples while nonetheless attaining very high likelihood. As
optimization proceeds,~$M_f$ also becomes increasingly ill-conditioned. This shows that likelihood
for generators need not correspond to intuitive notions of visual plausibility.

\section{Uncovering Latent Explanatory Factors}

\begin{figure}[t]
\centering
\subfloat[][Log spectra for CelebA models with~$n_f = 64$ and~$n_z$ varied.]{%
	\includegraphics[width=\textwidth]{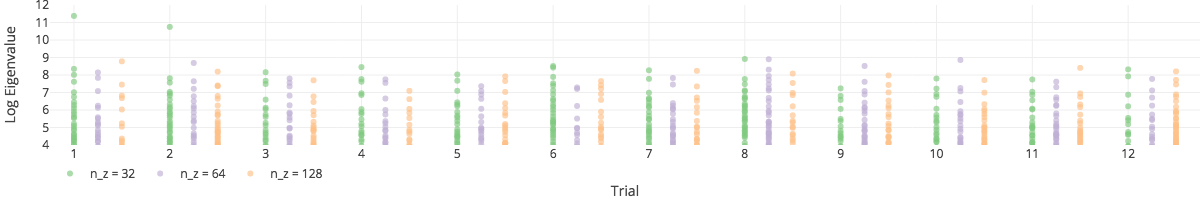}
} \\
\subfloat[][Log spectra for LSUN Bedroom models with~$n_f = 256$ and~$n_z$ varied.]{%
	\includegraphics[width=\textwidth]{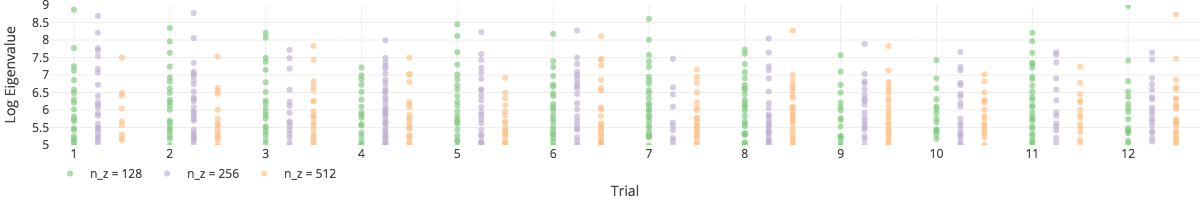}
} \\
\begin{tabular}[t]{c@{}c}%
\captionsetup{width=.45\linewidth,justification=centering}%
\subfloat[][CelebA ($n_z = 32, n_f = 64$), trials 10 and 11. Step~size:~0.40.]{%
	{\def\arraystretch{0}%
	\begin{tabular}[b]{c}%
	\includegraphics[width=0.47\textwidth]{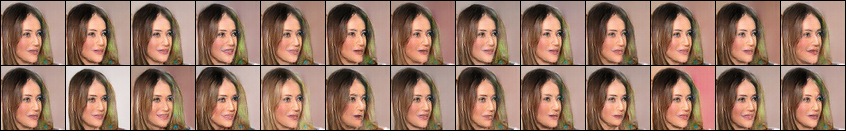} \\
	\includegraphics[width=0.47\textwidth]{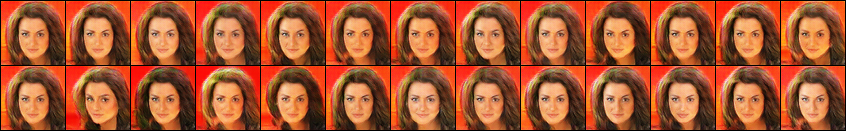}
	\end{tabular}}%
} &
\captionsetup{width=.45\linewidth,justification=centering}%
\subfloat[][LSUN Bedroom ($n_z = 128, n_f = 64$), trials 5 and 12. Step~size:~0.80.]{%
	{\def\arraystretch{0}%
	\begin{tabular}[b]{c}%
	\includegraphics[width=0.47\textwidth]{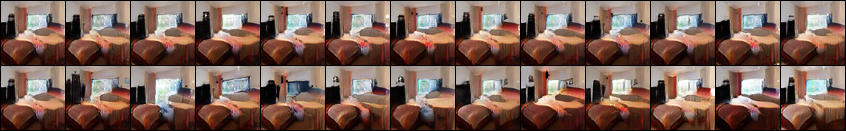} \\
	\includegraphics[width=0.47\textwidth]{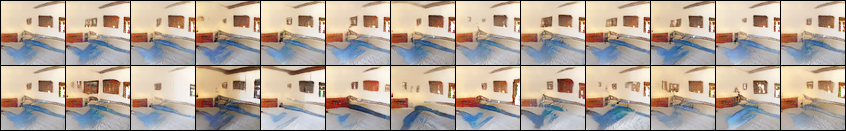}
	\end{tabular}}%
} \\
\captionsetup{width=.45\linewidth,justification=centering}%
\subfloat[][CelebA ($n_z = 64, n_f = 64$), trials 6 and 12. Step~size:~0.65.]{%
	{\def\arraystretch{0}%
	\begin{tabular}[b]{c}%
	\includegraphics[width=0.47\textwidth]{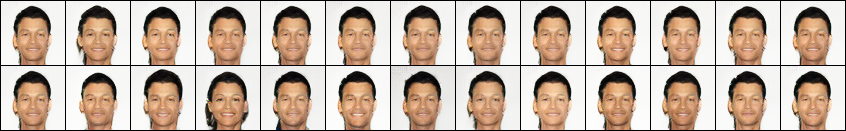} \\
	\includegraphics[width=0.47\textwidth]{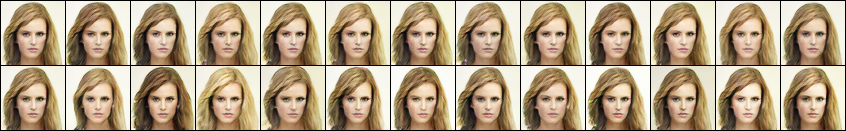}
	\end{tabular}}%
} &
\captionsetup{width=.45\linewidth,justification=centering}%
\subfloat[][LSUN Bedroom ($n_z = 256, n_f = 64$), trials 6 and 8. Step~size:~0.80.]{%
	{\def\arraystretch{0}%
	\begin{tabular}[b]{c}%
	\includegraphics[width=0.47\textwidth]{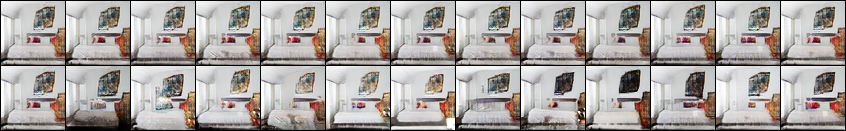} \\
	\includegraphics[width=0.47\textwidth]{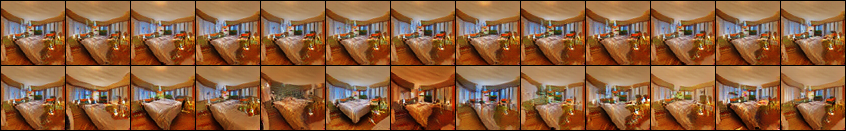} 
	\end{tabular}}%
} \\
\captionsetup{width=.45\linewidth,justification=centering}%
\subfloat[][Trials 7 and 9 ($n_z = 128, n_f = 64$).\\ Step~size:~0.80.]{%
	{\def\arraystretch{0}%
	\begin{tabular}[b]{c}%
	\includegraphics[width=0.47\textwidth]{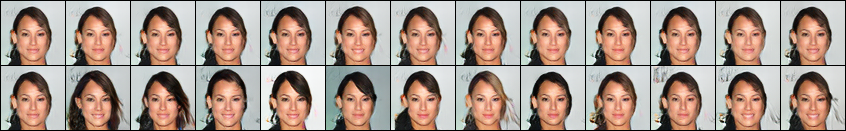} \\
	\includegraphics[width=0.47\textwidth]{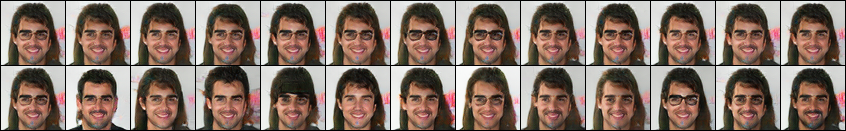}
	\end{tabular}}%
} &
\captionsetup{width=.45\linewidth,justification=centering}%
\subfloat[][Trials 8 and 10 ($n_z = 512, n_f = 64$).\\ Step~size:~0.80.]{%
	{\def\arraystretch{0}%
	\begin{tabular}[b]{c}%
	\includegraphics[width=0.47\textwidth]{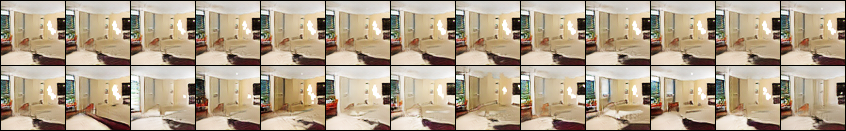} \\
	\includegraphics[width=0.47\textwidth]{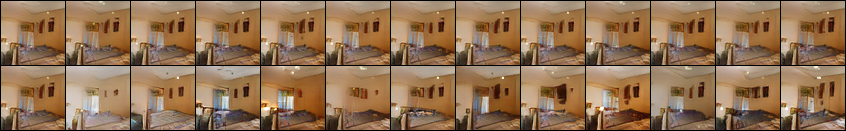}
	\end{tabular}}%
}%
\end{tabular}
\caption{Result of perturbing latent variables along the eigenvectors of~$M_f$. In~(a) and~(b), we
show the top eigenvalues of $\Sp(M_f)$ evaluated at~12 trial latent variables. Small eigenvalues are
not shown. The leftmost images of each grid both contain duplicates of the original, each
corresponding one of the~12 latent variables. The top row shows the effect of applying perturbations
along random directions, and the bottom row the result of applying perturbations with the same step
size along the eigenvectors.\label{fig:latent_perturbations}}
\end{figure}

The generator in GANs is well-known for organizing latent space such that semantic features can be
transferred by means of algebraic operations over latent variables~\citep{DCGAN}. This suggests the
existence of a systematic organization of latent space, but perhaps one that cannot be globally
characterized in terms of a handful of simple explanatory factors. We instead explore whether
\emph{local} changes in latent space can be characterized in this way. Since the metric~$M_f$
describes local change in the generator's output, it is natural to consider the effect of
perturbations along its eigenvectors. To this end, we fix~12 trial embeddings in latent space, and
compare the effect of perturbations along random directions to perturbations along these
eigenvectors. The random directions are obtained by sampling from a spherical normal distribution.
We show the results in Figure~\ref{fig:latent_perturbations}.

We can see that dominant eigenvalues, especially the principal eigenvalue, often result in the most
drastic changes. Furthermore, these changes are not only semantically meaningful, but also tend to
make modifications to distinct attributes of the image. To see this more clearly, we consider the
top two rows of Figure~\ref{fig:latent_perturbations}(g). Movement along the first two eigenvectors
changes hair length and facial orientation; movement along the third eigenvector decreases the
length of the bangs; movement along the fourth and fifth eigenvectors changes background color; and
movement along the sixth and seventh eigenvectors changes hair color.

\begin{figure}
\centering%
{\captionsetup{width=.23\linewidth,justification=centering}%
\subfloat[][CelebA \\ ($n_f = 64$, $n_z$ varied).]{%
	\includegraphics[width=0.25\textwidth]{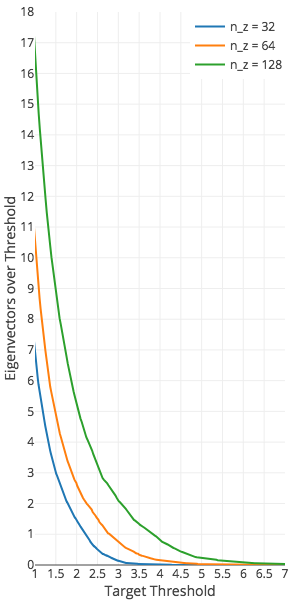}
}}%
{\captionsetup{width=.23\linewidth,justification=centering}%
\subfloat[][LSUN Bedroom \\ ($n_f = 64$, $n_z$ varied).]{%
	\includegraphics[width=0.25\textwidth]{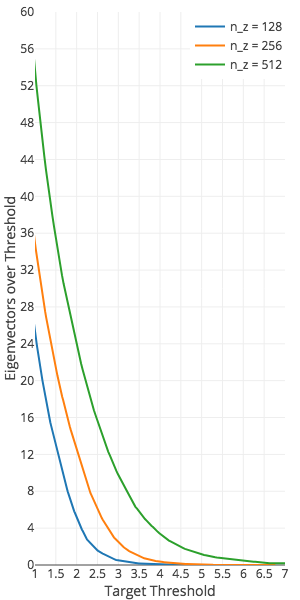}
}}%
{\captionsetup{width=.23\linewidth,justification=centering}%
\subfloat[][CelebA \\ ($n_z = 128$, $n_f$ varied).]{%
	\includegraphics[width=0.25\textwidth]{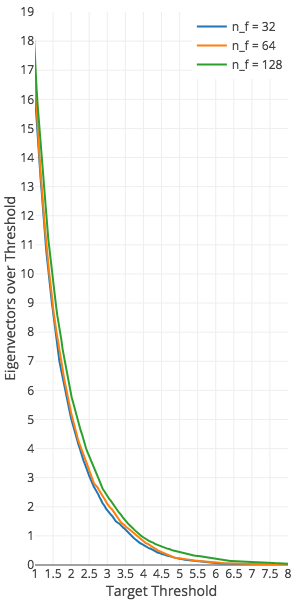}
}}%
{\captionsetup{width=.23\linewidth,justification=centering}%
\subfloat[][LSUN Bedroom \\ ($n_z = 256$, $n_f$ varied).]{%
	\includegraphics[width=0.25\textwidth]{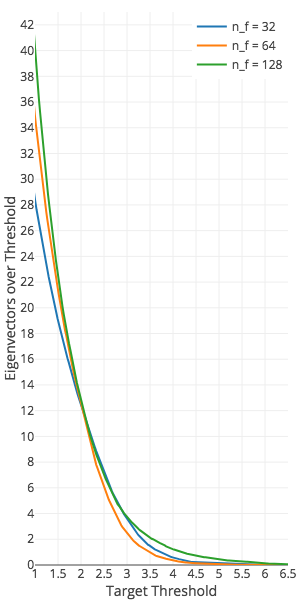}
}}

\subfloat[][Effect of varying~$n_z$ and~$n_f$ on~$\effFac(\tau = 1)$.]{\tiny%
\begin{tabular}{c NNN NNN}
\toprule
\multirow{2}{*}{Dataset} &%
\multicolumn{3}{c}{$\effFac(\tau = 1)$ for ${n_f}$ fixed, ${n_z}$ varied} &%
\multicolumn{3}{c}{$\effFac(\tau = 1)$ for ${n_z}$ fixed, ${n_f}$ varied} \\
\cmidrule(lr){2-4} \cmidrule(lr){5-7} &%
${(n_z / 2, n_f)}$ & ${(n_z, n_f)}$ & ${(2 n_z, n_f)}$ & ${(n_z, n_f / 2)}$ & ${(n_z, n_f)}$ & ${(n_z, 2 n_f)}$ \\
\midrule%
CelebA ($n_z = 64, n_f = 64$) &%
6.94921875 & 10.4921875 & 16.55078125 & 16.140625 & 16.55078125 & 17.28515625 \\
LSUN Bedroom ($n_z = 64, n_f = 128$) &%
25.2265625 & 34.9296875 & 53.6015625 & 28.3671875 & 34.9296875 & 40.04296875 \\
\bottomrule%
\end{tabular}}%
\caption{Effect of varying~$n_z$ and~$n_f$ on~$\effFac(\tau)$. Plots~(a)--(d) show~$\effFac$ as a
function of~$\tau$. Larger values for~$\effFac$ suggest increased utilization of latent space.
Table~(e) shows that doubling~$n_z$ roughly doubles~$\effFac$. On the other hand, doubling~$n_f$
does not result in noticeable change for CelebA, and only results in a modest increase in~$\effFac$
for LSUN Bedroom. The step sizes used for the perturbations are the same as those reported in
Figure~\ref{fig:latent_perturbations}.\label{fig:eff_fac}}
\end{figure}

Inspecting the two columns~(c), (e), (g), and~(d), (f), (h) in Figure~\ref{fig:latent_perturbations}
suggests that larger values of~$n_z$ may encourage the generator to capture more explanatory
factors, possibly at the price of decreased sample quality. We would like to explore the effect of
varying~$n_z$ and~$n_f$ on the number of such factors. To do this, we fix a sample of latent
variables~$S \coloneqq \{z_j\}_{j \in [1, M]} \sim P_Z$. For each~$z_j \in S$, we define
\[
	\delta(j, 0) \coloneqq \E_{\epsilon \sim N(0, I)} \norm{G(z_j + \alpha \epsilon) - G(z_j)}_2
	\quad\text{and}\quad
	\delta(j, i) \coloneqq \frac{\norm{G(z_j + \alpha v_i^{(j)}) - G(z_j)}_2}{\delta(j, 0)},
\]
for every eigenvector~$v_1^{(j)}, \ldots, v_{n_z}^{(j)}$ of~$M_f(z_j)$. The quantity~$\delta(j, i)$
measures the pixelwise change resulting from a perturbation along an eigenvector, relative to the
change we expect from a random perturbation. Finally, we define
\[
	\effFac(\tau) \coloneqq \frac{1}{M} \sum_{j \in [1, M]} \sum_{i \in [1, n_z]}
		\ind{\delta(j, i) > \tau},
\]
where~$\ind{\cdot}$ is the indicator function. This quantity measures the average number of
eigenvectors for which the relative change is greater than the threshold~$\tau$. As such, it can be
regarded as an effective measure of dimensionality for latent space. We explore the effect of
varying~$n_z$ and~$n_f$ on~$\effFac$ in Figure~\ref{fig:eff_fac}.

\section{Conclusion}

Current approaches for probabilistic modeling attempt to satisfy two goals that are fundamentally at
odds with one another: fulfillment of the two constraints described in Section~\ref{sec:intro}, and
model flexibility. In this work, we develop a computational tool that aims to expand the scope of
probabilistic modeling to functions that do not satisfy these constraints. We make the first steps
toward demonstrating feasibility of this approach by minimizing divergences in far greater
generality than what has been attempted before. Finally, we uncover surprising facts about the
organization of latent space for GANs that we hope will contribute to an improved understanding of
how effectively they capture underlying latent structure.

\bibliographystyle{plainnat}
\bibliography{main}

\end{document}